\documentclass[11pt,a4paper]{article}
\usepackage{authblk}
\usepackage[hyperref]{acl2020}
\usepackage{times}
\usepackage{latexsym}
\usepackage{multicol}  
\usepackage{multirow} 
\usepackage{url}
\usepackage{xcolor}
\usepackage{ulem}
\usepackage{adjustbox}
\usepackage{graphicx}

\usepackage{amsfonts}  
\usepackage{amssymb}
\usepackage{epstopdf}
\usepackage{enumitem}
\usepackage{mathtools} 
\usepackage{bbm}
\usepackage{url}
\usepackage{wrapfig}
\usepackage{appendix}
\usepackage{cases}
\usepackage{enumitem}
\usepackage{flexisym}

\usepackage{helvet}  
\usepackage{courier}  
\usepackage{url}  
\usepackage{graphicx}  
\usepackage[hang,flushmargin]{footmisc}
\usepackage{algorithm}  
\usepackage{algpseudocode}  
\usepackage{amsmath}
\usepackage{multicol}  
\usepackage{multirow} 

\usepackage{array}
\newcolumntype{L}[1]{>{\raggedright\let\newline\\\arraybackslash\hspace{0pt}}m{#1}}
\newcolumntype{C}[1]{>{\centering\let\newline\\\arraybackslash}m{#1}}
\newcolumntype{R}[1]{>{\raggedleft\let\newline\\\arraybackslash\hspace{0pt}}m{#1}}

\usepackage{cleveref}
\crefformat{section}{\S#2#1#3} 
\crefformat{subsection}{\S#2#1#3}
\crefformat{subsubsection}{\S#2#1#3}

\aclfinalcopy 

\title{Prototype-to-Style: Dialogue Generation \\with Style-Aware Editing on Retrieval Memory}

\author[1]{Yixuan Su}
\author[2]{Yan Wang}
\author[1]{Simon Baker}
\author[3]{Deng Cai}
\author[2]{Xiaojiang Liu}
\author[1]{Anna Korhonen}
\author[1]{Nigel Collier}
\affil[1]{University of Cambridge}
\affil[2]{Tencent AI Lab}
\affil[3]{The Chinese University of Hong Kong}
\affil[ ]{{\{ys484,sb895,alk23,nhc30\}@cam.ac.uk, thisisjcykcd@gmail.com}}
\affil[ ]{{\{brandenwang,kieranliu\}@tencent.com}}

\date{}

\begin{document}
\maketitle
\begin{abstract}

The ability of a dialog system to express prespecified language style during conversations has a direct, positive impact on its usability and on user satisfaction. We introduce a new prototype-to-style (PS) framework to tackle the challenge of stylistic dialogue generation. The framework uses an Information Retrieval (IR) system and extracts a response prototype from the retrieved response. A stylistic response generator then takes the prototype and the desired language style as model input to obtain a high-quality and stylistic response. To effectively train the proposed model, we propose a new style-aware learning objective as well as a de-noising learning strategy. Results on three benchmark datasets from two languages demonstrate that the proposed approach significantly outperforms existing baselines in both in-domain and cross-domain evaluations\footnote{All code and trained models will be made publicly available.}. 
\end{abstract}

\section{Introduction}

Most early research on dialogue response generation focused on generating grammatical and contextually relevant responses \cite{DBLP:conf/emnlp/RitterCD11,DBLP:journals/sigkdd/ChenLYT17,Martinovsky_2003.theerror}. While promising results have been demonstrated \cite{DBLP:conf/emnlp/WenGMRSUVY16,DBLP:conf/icassp/WangZTLL16}, syntactically coherent responses alone do not guarantee an engaging and attractive dialogue system. Expressing a unique and consistent speaking style has been shown to be crucial for increasing the user's engagement with dialogue systems \cite{DBLP:conf/cvpr/GanGHGD17}. There are various definitions of language style \cite{article_Roberts,bell_1984,bell1997towards,article_Niederhoffer,traugott_1975}. In this work, from a purely computational standpoint, we refer to language style as any characteristic style of expression. Hence, our work is in line with previous work on dialogue generation with emotion \cite{DBLP:conf/aaai/ZhouHZZL18,DBLP:conf/naacl/HuangZTD18,DBLP:conf/acl/WangZ18,DBLP:conf/aaai/Zhong0M19}; response attitude \cite{DBLP:journals/tacl/NiuB18}, and speaker personality \cite{DBLP:conf/acl/LiGBSGD16}. 

The aforementioned approaches explicitly incorporate the language style information into the model configuration either via embeddings or memory modules to control the process of response generation. In our replication experiments, we found that these approaches tend to overemphasise the importance of the language style. As a result, the generated responses tend to be generic and non-informative \cite{DBLP:conf/naacl/LiGBGD16}, but they do express a distinct style; e.g., they generate a generic response: \textit{``I am happy to hear that."} that conveys a `happy' emotion to different queries.


In this work, we propose a novel prototype-to-style (PS) framework to tackle the challenge of stylistic dialogue generation.
Our motivation is two-fold: (1) Human-written responses are informative and diverse, which could be leveraged as guidance for the generation model; (2) However, the retrieved response is not guaranteed to express the desired language style. Moreover, the quality of the retrieved response varies among different queries due to the instability of the IR system. Therefore, to transform the retrieved result into a relevant and stylistic response, an adequate editing process is necessary. 

\begin{figure}[tb] 
	\centering    
	\setlength{\abovecaptionskip}{3pt}
	\includegraphics[width=0.45\textwidth]{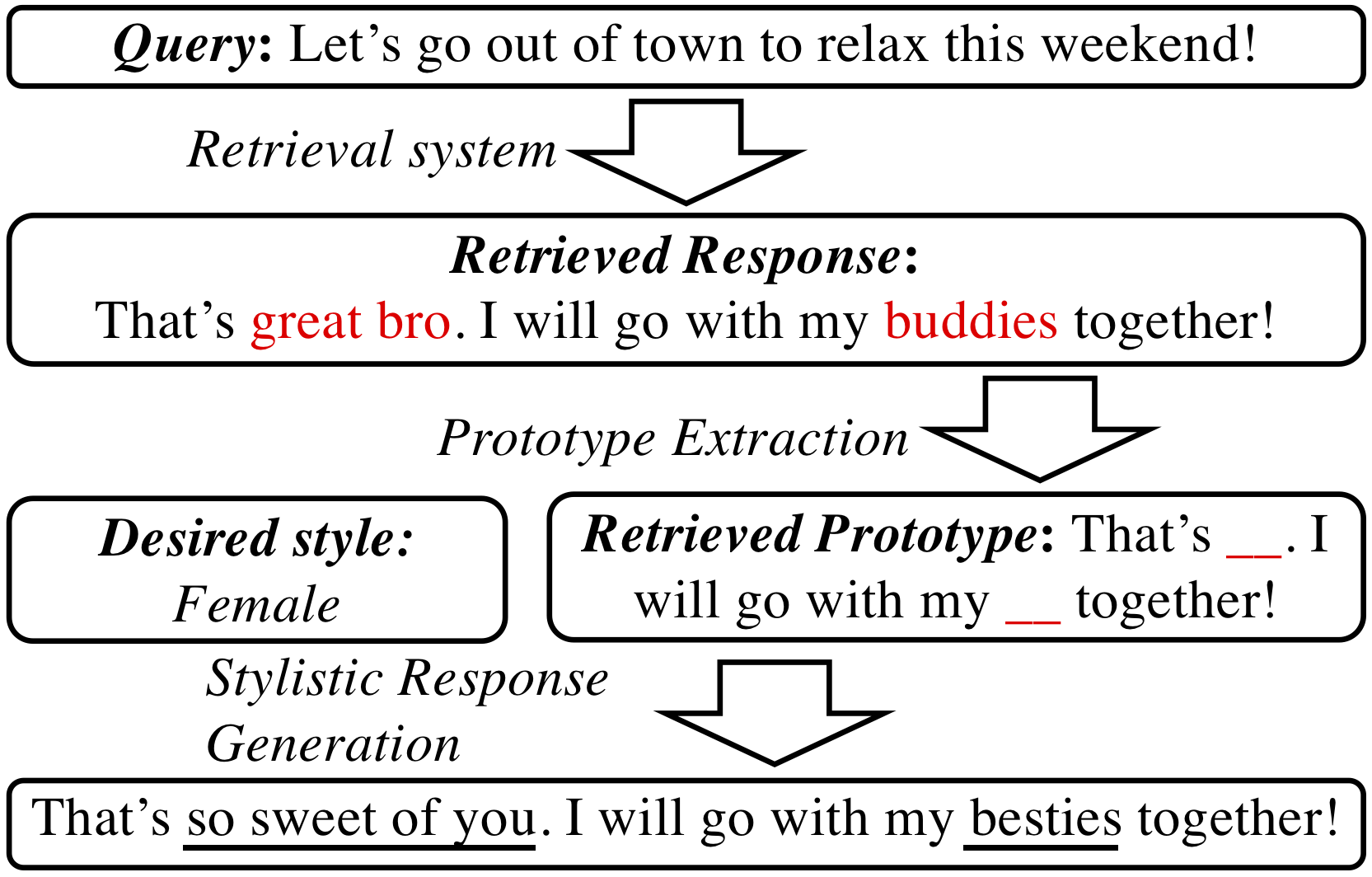}
	\caption{Prototype-to-Style Framework: It first constructs a neutral response prototype by \textit{masking} the stylistic words from the retrieved response. The stylistic response generator then takes the extracted prototype and the desired language style information to generate an adequate and stylistic response.}
	\label{fig:pipeline}
\end{figure}

An illustration of the proposed framework is shown in Figure \ref{fig:pipeline}, where a prototype is first extracted from the retrieved response. The stylistic response generator then takes the desired language style and the extracted prototype as additional input to obtain an adequate and stylistic response. The proposed stylistic response generator mainly inherits from the GPT-2 model \cite{Radford2019LanguageMA} which is pre-trained with a large unlabeled text corpus. However, the GPT-2 model does not naturally fit the task of dialogue generation. To this end, we design various adaptations to the model architecture to extend the GPT-2 model to address the task of dialogue generation. Furthermore, in order to control the style of the generated responses, we train the model with a novel style-aware maximum likelihood estimation (MLE) objective that encodes additional style knowledge into the model's parameters. Finally, to mitigate the possible effect that the retrieved response containing irrelevant and inappropriate information with respect to the input query, we adopt a de-noising learning strategy \cite{DBLP:conf/nips/JainS08,DBLP:conf/cvpr/KrullBJ19} to prevent the model from uncritically copying the prototype. 


To fully evaluate the proposed approach, we conduct extensive experiments on three benchmark datasets. Results of both human and automatic evaluation show that the proposed approach significantly outperforms several strong baselines. In addition, we also conduct an extensive cross-domain experiment to demonstrate that the proposed approach is more robust than such baselines.



It should be noted that stylistic dialogue generation is different from the task of text style transfer. Text style transfer aims to rewrite the input sentences such that they possess certain language styles, while rigorously preserving their semantic meaning \cite{DBLP:journals/corr/abs-1901-11333}. On the other hand, stylistic dialogue generation does not aim at preserving the semantic meaning of the input sentences. Instead, it aims at generating sentences that are adequate and relevant responses to the input sentences, while expressing the prespecified language styles. 

In summary, the contributions of this work are: (1) We propose a novel framework that tackles the challenge of stylistic dialogue generation by leveraging useful information contained in the retrieved responses; (2) We propose a new stylistic response generator by making proper adaptations to a large-scale pre-trained language model. We train our model with a new style-aware learning objective in a de-noising manner. Experiments show that the proposed model outperforms many strong baselines on three benchmark datasets on both in-domain and cross-domain evaluations.


\section{Related Work}
We summarize three categories of relevant work in the following.
\paragraph{Text Style Transfer:}
The task of text style transfer aims to transfer the style contained in a sentence while preserving its meaning. \citet{DBLP:conf/naacl/LiJHL18} proposed a DRG framework to tackle this task with the help of external knowledge. Recently, based on the pre-trained language model, \citet{DBLP:journals/corr/abs-1908-09368} further improved the system performance under the same DRG framework.

\paragraph{Retrieval Guided Dialogue Generation:}
Many prior works \cite{DBLP:conf/ijcai/SongLNZZY18,DBLP:conf/acl/ZhuCZWL19,DBLP:conf/aaai/0006WHWL019,DBLP:conf/naacl/CaiWBTLLS19} proposed to leverage information from the retrieved responses to improve the system performance on non-task oriented dialogue generation. It should be noted that all these approaches aim to improve the content quality of the generated responses but do not take the style aspect into consideration.

\paragraph{Stylistic Dialogue Generation:} 
Extensive research has tried to tackle the task of stylistic dialogue generation. \citet{DBLP:conf/acl/LiGBSGD16} proposed to represent the user's personality with embeddings and incorporated them into the decoder structure to control the response generation process. \citet{DBLP:journals/tacl/NiuB18} used reinforcement learning to train the generation model via the interaction with a pre-trained classifier to generate responses with specified attitude. \citet{DBLP:conf/aaai/ZhouHZZL18,DBLP:conf/naacl/HuangZTD18,DBLP:conf/acl/WangZ18,DBLP:conf/aaai/Zhong0M19} incorporated external knowledge into the model architecture either via embeddings or internal and external memory modules, such that during the generation process, emotion-based styles can be dynamically controlled. \citet{DBLP:conf/emnlp/GaoZLGBGD19} proposed to use a shared latent space for stylistic dialogue generation. 

\begin{figure*}[tb] 
	\centering    
	\setlength{\abovecaptionskip}{3pt}
	\includegraphics[width=0.85\textwidth]{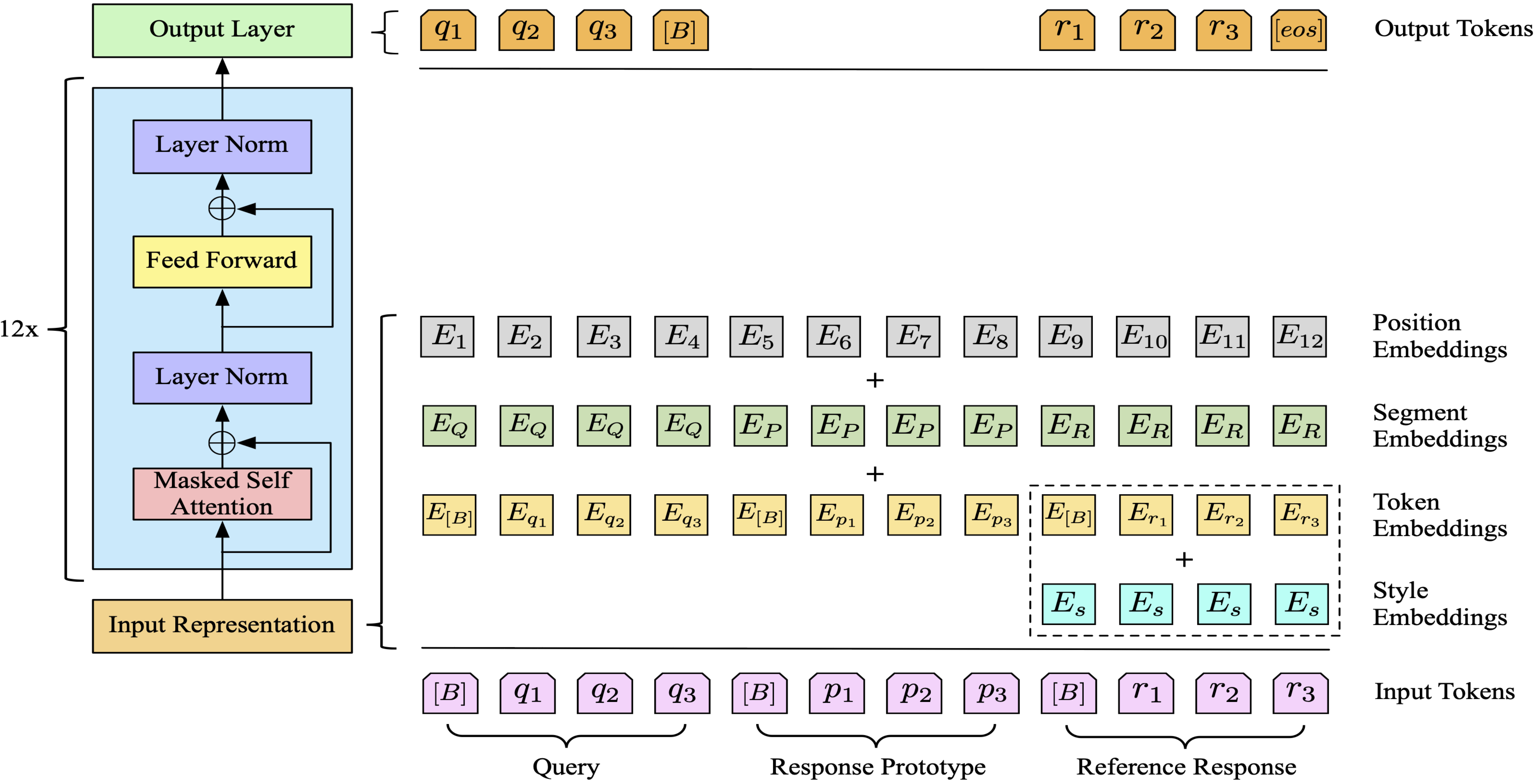}
	\caption{Illustration of the proposed Stylistic Response Generator: The input representation is constructed by adding up four different level embeddings. By specifying different style embeddings, the model can generate responses with different language styles.}
	\label{fig:model}
\end{figure*}

\section{Methodology}
The proposed framework leverages the results acquired from an IR system, A major challenge is that the retrieved response is not guaranteed to express the desired language style. At the first step, a neutral response prototype is extracted by masking all stylistic words contained in the retrieved response. A stylistic response generator then takes the desired language style and the extracted prototype as additional input to generate an adequate and stylistic response to the input query. To better emphasize the generation of stylistic expressions, we propose a style-aware learning objective. Finally, to prevent the model from learning to uncritically copy the prototype, we adopt a de-noising learning strategy \cite{DBLP:conf/nips/JainS08,DBLP:conf/cvpr/KrullBJ19} to train the generator. 

\subsection{Prototype Extraction}
\label{SV}
The response prototype is constructed from the retrieved response by masking the stylistic words. To determine whether a word is stylistic, we use the pointwise mutual information (PMI) \cite{church1990word} metric. The relevance between the word $x$ and the style $s$ is measured as \begin{equation}
    \nonumber
    \textup{PMI}(x;s) = \log\frac{p(x,s)}{p(x)p(s)},
    \label{eq:pmi}
\end{equation}
where $p(x, s)$ is the frequency that the word $x$ appears in a response with style $s$ in the training corpus. And a word $x$ is stylistic given the style $s$ if $\textup{PMI}(x,s)\geq t_s$. In our experiments, we empirically set $t_s$ as $t_s = \frac{3}{4}\times\max_{v\in\mathcal{V}}\textup{PMI}(v; s)$, where $\mathcal{V}$ is the vocabulary set of the training corpus. Given the set of all possible language styles $\mathcal{S}$, the stylistic vocabulary $\mathcal{SV}$ is defined as all words that express any style $s\in \mathcal{S}$. An example is provided in Figure \ref{fig:pipeline} where the prototype:
\textit{``That's \_ . I will go with my \_ together !''} is extracted from the retrieved response by masking the stylistic words \textit{great}, \textit{bro} and \textit{buddies}.

\subsection{Stylistic Response Generator}
The proposed Stylistic Response Generator inherits from the GPT-2 \cite{Radford2019LanguageMA} model which consists of a 12-layer decoder-only Transformer \cite{DBLP:conf/nips/VaswaniSPUJGKP17}. To make use of the GPT-2 model, the input tokens must be a consecutive natural sequence (e.g. sentence, document). Based on the input sequence, the input representation is constructed by adding up the token embeddings and the corresponding position embeddings. 

To achieve the goal of adapting the GPT-2 model under the proposed PS framework, we first make modifications to the form of the input sequence. As shown in Figure \ref{fig:model}, we construct the input sequence as the concatenation of the input query, the response prototype and the reference response. Then we introduce a special token $[B]$ to indicate the boundary between these three parts. To further ensure the model can identify the different parts of the input sequence, we introduce a new segment level input which consists of three learnable segment embeddings $E_Q$, $E_P$ and $E_R$ to indicate the positions of the input query, the response prototype and the response history. 

To control the language style of the generated response, we propose to incorporate learnable style embeddings into the input representation. Specifically, we add the style embeddings\footnote{Each style embedding corresponds to one specific language style; e.g. if we consider three different gender styles, the number of different style embeddings is $3$.} to the entire part of the response history. This way, the model is constantly aware of the desired language style through the entire generation process. 


\subsection{Learning}
\subsubsection{Style-Aware Learning Objective}
We propose to use a new style-aware learning objective to train the stylistic response generator. Consider a training instance consists of the input query ${\bf X} = (x_1, ..., x_N)$, the reference response ${\bf Y} = (y_1, ..., y_T)$, the reference language style $s$ and the response prototype ${\bf C} = (c_1, ..., c_T)$, the proposed objective is defined as
\begin{equation} 
\nonumber
\begin{split}
    &L_{\textup{S-MLE}}(\theta) =\\
    &-\sum_{i=1}^{T}\log p_{\theta}(y_i|y_1, ..., y_{i-1}; {\bf X}, {\bf C}, s)\cdot f(y_i)
\end{split}
\label{eq:se-mle}
\end{equation}
$$ 
    f(y_i) =
      \begin{cases} 
      1 + \alpha & \mbox{if } y_i\in \mathcal{SV}\\
      1 & \textup{otherwise},
\end{cases} $$
where $\theta$ are the model parameters and $\mathcal{SV}$ is the stylistic vocabulary introduced in \cref{SV}. By increasing $\alpha$, the proposed objective encodes more knowledge about stylistic expressions into the model parameters.

We find that including the language model as an auxiliary objective in addition to the supervised style-aware learning objective helps to improve generalization as well as accelerate convergence. This observation is in line with \citet{DBLP:conf/acl/Rei17,OpenAI_GPT}. In this work, the language model objective is defined as the reconstruction loss of the input query based on itself:
\begin{equation} 
\nonumber
\begin{split}
    L_{\textup{LM}}(\theta) &= -\log p_{\theta}({\bf X})\\
    &= -\sum_{j=2}^{N}\log p_{\theta}(x_j|x_1, ..., x_{j-1}).
\end{split}
\label{eq:lm}
\end{equation}
The final learning objective is then defined as
\begin{equation}
    \nonumber
    L(\theta) = L_{\textup{S-MLE}}(\theta) + \beta L_{\textup{LM}}(\theta),
    \label{eq:lo}
\end{equation}
where $\beta$ regulates the importance of the auxiliary objective\footnote{The $\alpha$ in $L_{\textup{S-MLE}}(\theta)$ is set to be 0.2 and the $\beta$ is set to be 1.0}. 

\begin{figure}[tb] 
	\centering    
	\setlength{\abovecaptionskip}{3pt}
	\includegraphics[width=0.42\textwidth]{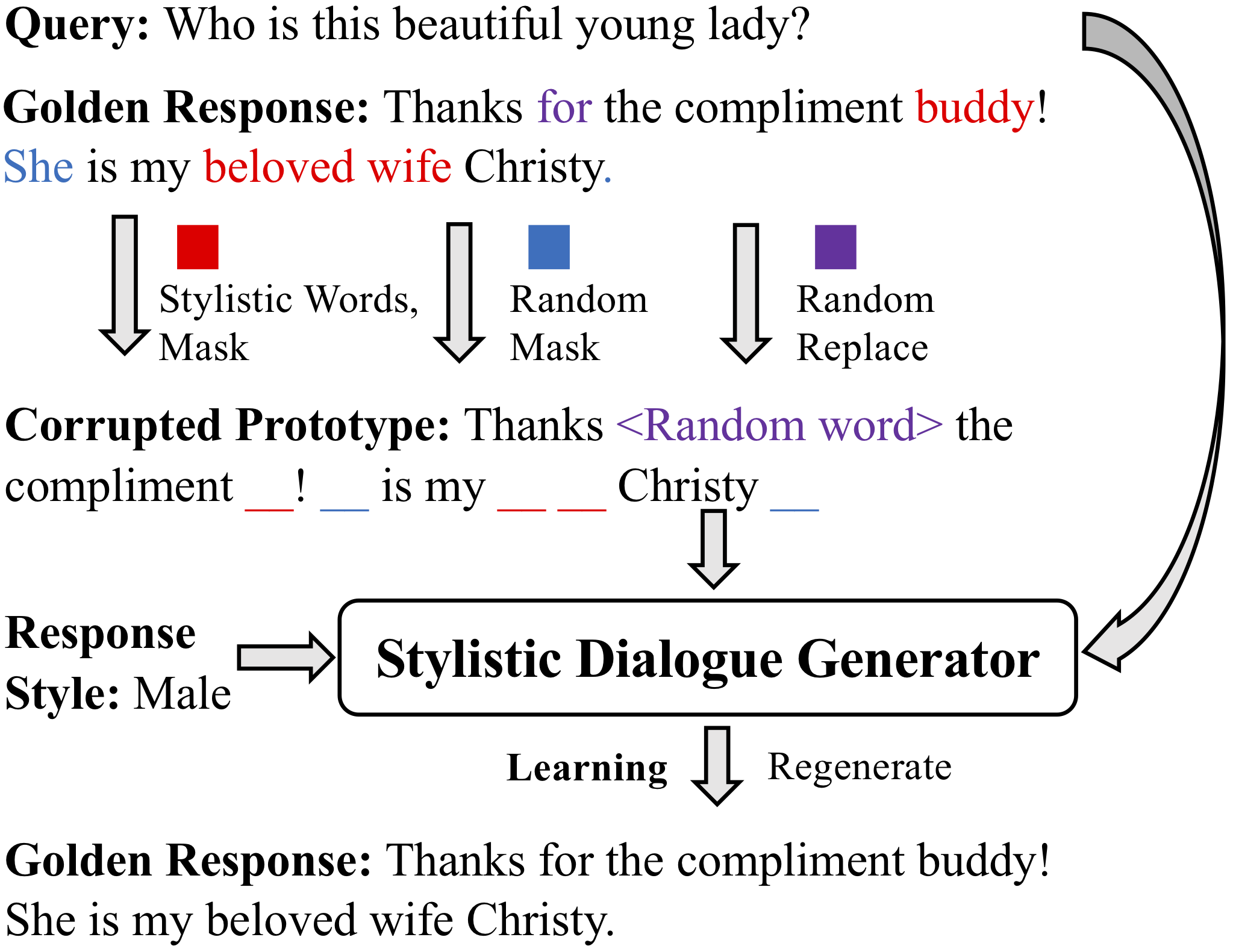}
	\caption{Illustration of de-noising training strategy.}
	\label{fig:denoising_training}
\end{figure}

\subsubsection{De-noising Training}
We use a de-noising training strategy similar to \newcite{DBLP:conf/nips/JainS08, DBLP:conf/cvpr/KrullBJ19} for training data construction, as shown in Figure \ref{fig:denoising_training}. Specifically, during training, the response prototype is extracted from the reference response by the following steps. First, we mask all the stylistic words in the reference response. Second, we randomly select some words (40\%) and replace it with a special token [MASK] or a random word drawn from the vocabulary.

The second step is necessary otherwise the model will learn to generate a response by uncritically copying the response prototype, since the prototype after the first step is always an integral part of the golden response. This copy mechanism is undesirable since during testing the retrieved response is likely to contain information that is irrelevant to the input query. Thus, we deliberately train the response generator with noisy input to let the model learn to filter out the inappropriate information contained in the response prototype.

\section{Datasets}
We conduct extensive experiments on three dialogue datasets: gender-specific (Chinese) dataset, emotion-specific (Chinese) dataset, and sentiment-specific (English) dataset. For each dataset, we randomly select 200 instances as a held-out test set for evaluation. 


\subsection{Gender-Specific Dialogue Dataset}
We use a publicly available gender-specific dialogue dataset \cite{IGRL}. In this dataset, each response contains one specific gender preference including  \textit{Female}, \textit{Male} and \textit{Neutral}.

\subsection{Emotion-Specific Dialogue Dataset}
We use a publicly available emotion-specific dataset \cite{DBLP:conf/aaai/ZhouHZZL18} which contains responses with $6$ different emotions including \textit{Like}, \textit{Disgust}, \textit{Happy}, \textit{Anger}, \textit{Sad} and \textit{Other}. 

\begin{table}[tb]
    \footnotesize
	\setlength{\abovecaptionskip}{3pt}
	\renewcommand\arraystretch{1.1}
	\centering  
    \begin{center}
    \scalebox{0.9}{
	\begin{tabular}{|c|c|c|c|}
	    \hline
		Queries&\multicolumn{2}{|c|}{26,265,224}&Percentage(\%)\\
		\hline
        \multirow{3}{*}{Responses}&Positive&4,275,978&16.28\%\\\cline{2-4}
        &Negative&6,282,641&23.92\%\\\cline{2-4}
        &Neutral&15,706,605&59.80\%\\
        \hline
	\end{tabular}}
	\end{center}
	\caption{Data Statistic of Sentiment-Specific Dataset}
	\label{tb:senti_stat}
\end{table}

\subsection{Sentiment-Specific Dialogue Dataset}
To construct this dataset, we first build a classifier on the basis of BERT \cite{DBLP:conf/naacl/DevlinCLT19} and finetuned it on the the SemEval-2017 Subtask A dataset \cite{rosenthal-etal-2017-semeval}. This dataset consists of twitter instances with different sentiments including \textit{Positive}, \textit{Negative} and \textit{Neutral}. 

The sentiment classifier attains 81.4\% classification accuracy which is further used to annotate the OpenSubtitles dataset \cite{DBLP:conf/lrec/LisonT16}. The data statistic of the resulting sentiment-specific dialogue dataset is shown in Table \ref{tb:senti_stat}.
 
\section{Experiments}
\subsection{Pretraining and Implementation Details}
As there is no off-the-shelf pre-trained word-level language model in Chinese, we manually pre-trained one. The corpus collection and model pre-training details are presented in the supplementary material. For the English pre-trained language model, we use the PyTorch adaptation released by the HuggingFace team\footnote{https://github.com/huggingface/pytorch-openai-transformer-lm}.

To optimize the model, we use the Adam optimizer \cite{DBLP:journals/corr/KingmaB14} with a batch size of 64 and learning rate of 2e-5. During inference, the retrieval system is built from the training corpus, and the retrieved responses are selected using the Jaccard similarity \cite{lipkus1999proof} between queries.

During the inference stage, we retrieve the candidates from the training set. Specifically, we employ Jacquard Similarity to calculate the similarity between the input query q and queries in training set and find the most similar query q$^\prime$. Then we directly adopt the response of the retrieved query q$^\prime$ to construct the response prototype.

\begin{table*}[tb]
	\setlength{\abovecaptionskip}{3pt}
	\renewcommand\arraystretch{1.1}
	\centering  
    \begin{center}
    \scalebox{0.63}{
	\begin{tabular}{|c|c|c|c|c|c|c|c|c|c|c|}
		\hline
		\multirow{ 2}{*}{Style}&\multirow{ 2}{*}{Metrics}&\multicolumn{4}{|c|}{Generative}&\multicolumn{3}{|c|}{Retrieval-Based}&\multicolumn{2}{|c|}{Ours}\\
		\cline{3-11}
		&&Seq2seq&GPT2-FT&Speaker&ECM&SR&RST&RRe&PS w/o R&PS\\\hline
		\hline
        \multirow{3}{*}{Male}& Quality$\uparrow$&2.97&3.33&2.49&2.56&2.58&2.15&2.78&2.94&\textbf{3.48}\\
        &Style Expression$\uparrow$&2.93&2.99&3.51&3.60&2.98&3.21&3.01&3.36&\textbf{3.75}\\
        &Ranking$\downarrow$&3.04&2.71&3.42&3.15&3.89&4.01&3.43&2.34&\textbf{1.56}\\
        \hline
        \multirow{3}{*}{Female}& Quality$\uparrow$&2.97&3.31&2.86&2.81&2.60&2.16&3.11&3.01&\textbf{3.42}\\
        &Style Expression$\uparrow$&3.07&3.02&3.01&3.09&3.02&3.14&3.09&3.49&\textbf{3.64}\\
        &Ranking$\downarrow$&2.94&2.62&3.18&3.20&3.66&3.86&2.89&2.28&\textbf{1.52}\\
        \hline
        \multirow{5}{*}{Overall}& Quality$\uparrow$&2.98&3.32&2.68&2.67&2.59&2.14&2.94&2.98&\textbf{3.45}\\
        &Style Expression$\uparrow$&3.00&3.05&3.26&3.35&3.03&3.17&3.01&3.43&\textbf{3.69}\\
        &Ranking$\downarrow$&2.99&2.66&3.30&3.17&3.78&3.94&3.16&2.31&\textbf{1.54}\\
        \cline{2-11}
        &Distinct-1(\%)$\uparrow$&27.64&36.42&26.15&12.45&37.62&33.12&\textbf{48.52}${\dagger}$&29.98&\textbf{40.88}\\
        &Distinct-2(\%)$\uparrow$&72.33&74.30&50.40&31.64&84.33&85.63&\textbf{94.11}${\dagger}$&78.54&\textbf{90.82}\\
        \hline
	\end{tabular}}
	\end{center}
	\caption{Evaluation Results on Gender-Specific Dialogue Generation: $\uparrow$ means the higher the better and $\downarrow$ means the lower the better, bold font denotes the best scores for each metric. Sign tests on evaluation scores show that the proposed model significantly outperforms other models with p-value $<$ 0.05 with the only exception marked by ${\dagger}$.}
	\label{tb:gender-result}
\end{table*}

\begin{table*}[tb]
	\setlength{\abovecaptionskip}{3pt}
	\renewcommand\arraystretch{1.1}
	\centering  
    \begin{center}
    \scalebox{0.63}{
	\begin{tabular}{|c|c|c|c|c|c|c|c|c|c|c|}
		\hline
		\multirow{2}{*}{Style}&\multirow{ 2}{*}{Metrics}&\multicolumn{4}{|c|}{Generative}&\multicolumn{3}{|c|}{Retrieval-Based}&\multicolumn{2}{|c|}{Ours}\\
		\cline{3-11}
		&&Seq2seq&GPT2-FT&Speaker&ECM&SR&RST&RRe&PS w/o R&PS\\\hline
		\hline
        \multirow{3}{*}{Like}& Quality$\uparrow$&3.06&3.48&2.62&2.61&2.49&2.25&2.61&3.49&\textbf{3.62}\\
        &Style Expression$\uparrow$&3.01&3.05&3.95${\dagger}$&\textbf{4.38}${\dagger}$&2.99&3.25&2.83&3.93${\dagger}$&3.77\\
        &Ranking$\downarrow$&3.77&3.53&3.47&2.95&4.71&4.43&4.56&2.11&\textbf{1.86}\\
        \hline
        \multirow{3}{*}{Disgust}& Quality$\uparrow$&3.03&\textbf{3.47}${\dagger}$&2.07&1.99&2.45&2.34&2.58&3.27&\textbf{3.41}\\
        &Style Expression$\uparrow$&2.53&2.68&\textbf{4.06}${\dagger}$&3.97${\dagger}$&2.85&3.17&2.99&3.39&3.61\\
        &Ranking$\downarrow$&3.95&3.51&3.79&3.97&4.45&3.95&4.11&2.15&\textbf{1.85}\\
        \hline
        \multirow{3}{*}{Happy}& Quality$\uparrow$&3.03&3.48&2.06&2.46&2.51&2.43&2.69&3.51&\textbf{3.68}\\
        &Style Expression$\uparrow$&4.06&3.49&4.83${\dagger}$&\textbf{4.94}${\dagger}$&3.09&4.73${\dagger}$&2.91&4.81${\dagger}$&4.59\\
        &Ranking$\downarrow$&3.44&3.91&4.19&3.32&5.51&3.31&5.35&1.85&\textbf{1.62}\\
        \hline
        \multirow{3}{*}{Anger}& Quality$\uparrow$&2.98&\textbf{3.43}${\dagger}$&1.94&1.95&2.41&2.31&2.66&3.01&\textbf{3.37}\\
        &Style Expression$\uparrow$&1.76&2.35&3.93${\dagger}$&\textbf{4.02}${\dagger}$&2.77&3.46&2.94&3.82${\dagger}$&3.83\\
        &Ranking$\downarrow$&5.27&4.05&4.21&3.97&4.71&3.75&4.05&2.20&\textbf{1.78}\\
        \hline
        \multirow{3}{*}{Sad}& Quality$\uparrow$&2.95&\textbf{3.44}${\dagger}$&2.14&2.09&2.37&2.30&2.59&3.12&\textbf{3.42}\\
        &Style Expression$\uparrow$&1.83&2.36&3.64${\dagger}$&3.43&2.77&3.24&2.92&\textbf{3.68}${\dagger}$&3.58\\
        &Ranking$\downarrow$&5.01&3.81&4.01&4.18&4.47&3.85&3.97&1.98&\textbf{1.81}\\
        \hline
        \multirow{5}{*}{Overall}& Quality$\uparrow$&3.01&\textbf{3.46}${\dagger}$&2.16&2.22&2.44&2.33&2.63&3.31&\textbf{3.46}\\
        &Style Expression$\uparrow$&2.64&2.79&4.08${\dagger}$&\textbf{4.14}${\dagger}$&2.89&3.57&2.92&3.93${\dagger}$&3.85\\
        &Ranking$\downarrow$&4.29&3.74&3.94&3.68&4.77&3.86&4.41&2.06&\textbf{1.78}\\
        \cline{2-11}
        &Distinct-1(\%)$\uparrow$&13.61&19.37&8.99&5.85&31.71&30.39&\textbf{44.67}${\dagger}$&22.55&\textbf{36.47}\\
        &Distinct-2(\%)$\uparrow$&34.03&59.65&21.08&15.68&78.91&82.41&\textbf{92.58}${\dagger}$&69.29&\textbf{87.48}\\
        \hline
	\end{tabular}}
	\end{center}
    \caption{Evaluation Results on Emotional-Specific Dialogue Generation}
	\label{tb:emotion-result}
\end{table*}

\begin{table*}[tb]
	\setlength{\abovecaptionskip}{3pt}
	\renewcommand\arraystretch{1.1}
	\centering  
    \begin{center}
    \scalebox{0.63}{
	\begin{tabular}{|c|c|c|c|c|c|c|c|c|c|c|}
		\hline
		\multirow{ 2}{*}{Style}&\multirow{ 2}{*}{Metrics}&\multicolumn{4}{|c|}{Generative}&\multicolumn{3}{|c|}{Retrieval-Based}&\multicolumn{2}{|c|}{Ours}\\
		\cline{3-11}
		&&Seq2seq&GPT2-FT&Speaker&ECM&SR&RST&RRe&PS w/o R&PS\\\hline
		\hline
        \multirow{3}{*}{Positive}& Quality$\uparrow$&2.63&2.97&2.72&2.72&1.90&2.42&2.49&2.93&\textbf{3.28}\\
        &Style Expression$\uparrow$&2.52&2.55&3.51&\textbf{3.89}${\dagger}$&2.72&2.96&2.70&3.44&3.76\\
        &Ranking$\downarrow$&4.39&4.05&3.10&2.38&4.71&4.10&4.12&2.61&\textbf{1.79}\\
        \hline
        \multirow{3}{*}{Negative}& Quality$\uparrow$&2.69&2.96&2.99&2.56&1.82&2.26&2.64&2.80&\textbf{3.20}\\
        &Style Expression$\uparrow$&3.15&3.09&\textbf{3.62}${\dagger}$&3.47&2.71&3.18&2.82&3.42&\textbf{3.63}\\
        &Ranking$\downarrow$&3.62&3.68&3.48&3.04&4.81&4.00&3.80&2.78&\textbf{2.39}\\
        \hline
        \multirow{5}{*}{Overall}& Quality$\uparrow$&2.66&2.97&2.86&2.64&1.86&2.34&2.57&2.87&\textbf{3.24}\\
        &Style Expression$\uparrow$&2.83&2.82&3.57&\textbf{3.68}${\dagger}$&2.72&3.07&2.76&3.43&\textbf{3.70}\\
        &Ranking$\downarrow$&4.00&3.85&2.79&2.71&4.76&4.05&3.96&2.69&\textbf{2.09}\\
        \cline{2-11}
        &Distinct-1(\%)$\uparrow$&24.65&29.92&23.61&14.22&30.06&40.13&\textbf{49.94}${\dagger}$&32.29&\textbf{44.70}\\
        &Distinct-2(\%)$\uparrow$&48.74&56.27&43.11&23.72&75.73&71.73&\textbf{91.59}${\dagger}$&68.35&\textbf{87.15}\\
        \hline
	\end{tabular}}
	\end{center}
	\caption{Evaluation Results on Sentiment-Specific Dialogue Generation}
	\label{tb:sentiment-result}
\end{table*}

\subsection{Model Comparison}
We compare the proposed approach with several competitive baselines that can be categorized into two classes: generative approaches and retrieval-based approaches.

\subsubsection{Generative Approaches}
\paragraph{Seq2seq:} Standard sequence-to-sequence model with attention mechanism \cite{DBLP:journals/corr/BahdanauCB14,DBLP:conf/emnlp/LuongPM15}.
\paragraph{GPT2-FT:} To examine the effect of leveraging the pre-trained language model for the task of dialogue generation, we directly fine-tune the GPT-2 model on the dialogue data without any designed adaptations.
\paragraph{Speaker:} Model proposed by \citet{DBLP:conf/acl/LiGBSGD16} which incorporates distributed style embeddings into the structure of decoding cells to control the generation process.
\paragraph{ECM:} Model proposed by \citet{DBLP:conf/aaai/ZhouHZZL18} which uses memory modules to control the stylistic expressions in the generated responses.

\subsubsection{Retrieval-Based Approaches}
\paragraph{Skeleton-to-Response (SR):} Model proposed by \citet{DBLP:conf/naacl/CaiWBTLLS19} which modifies the retrieved response based on the lexical difference between the input and the retrieved query. This approach does not take the style aspect into consideration.
\paragraph{Retrieval + Style Transfer (RST):} For this approach, we apply the state-of-the-art style transfer \cite{DBLP:journals/corr/abs-1908-09368} model on the retrieved response. This approach does not consider the input query information during the transfer process.
\paragraph{Retrieval + Reranking (RRe):} Given the input query, a style classifier is used to rerank the top 10 retrieved responses. The response with the highest score on the desired style is selected. 

\subsubsection{Ablation Study}
\paragraph{PS:} The full model proposed in this work.
\paragraph{PS w/o R:} In the ablated model, we examine how the retrieved prototype effects our model's performance. To this end, we remove the response prototype from the input representation. 

\subsection{Evaluation Metrics}
The quality of dialogue responses is known to be difficult to measure automatically \cite{DBLP:journals/corr/abs-1905-04071}; we therefore rely on human evaluation. To evaluate the responses, we hire five annotators from a commercial annotation company. To prevent introducing potential bias to the annotators, all results are randomly shuffled before being evaluated. All results are evaluated by the annotators following the metrics below.

\paragraph{Quality:} This metric evaluates the content quality of the generated responses. The annotators are asked to give a score within $5$-point scale where $5$ means \textit{perfectly human-like response} (relevant, fluent and informative), $3$ means marginally acceptable and $1$ means \textit{unreadable and impossible to understand}.

\paragraph{Style Expression:} This metric measures how well the generated responses express the desired style. The annotators give a score ranging from $1$ to $5$ to this metric, where $5$ means \textit{very strong style}, $3$ means \textit{no obvious style} and $1$ means \textit{very conflicted style}. The style conflict means the generated style is conflicted to the desired one (e.g. female to male, positive to negative emotion).

\paragraph{Ranking:} The annotators are further asked to jointly evaluate the content quality and the style expression of the generated responses from different approaches. Then the annotators give a ranking to each result where top $1$ means the best\footnote{The same ranking is allowed to be assigned to results from different approaches if they have the same overall quality.}.



\subsection{Main Results}
\label{sec:main_result}
Both human and automatic evaluation results on the three benchmark datasets are shown in Table \ref{tb:gender-result}, \ref{tb:emotion-result} and \ref{tb:sentiment-result}. For each dataset, we present results on individual styles as well as the overall results. 

We observe that the proposed model achieves the top performance results on most of the metrics. It generates responses with both intense style and high response quality.
In addition, we also measure the diversity of the generated responses with two automatic metrics: Distinct-1 and Distinct-2 \cite{DBLP:conf/acl/LiGBSGD16}. The results show that the proposed model achieves the closest performance to that of the RRe approach whose responses are all written by human. On the ranking metric which jointly evaluates the content quality and the style expression, the proposed model outperforms other approaches by a substantial margin.




From the results in Table \ref{tb:emotion-result} and \ref{tb:sentiment-result}, we can observe that ECM obtains the highest style expression scores on the emotion and sentiment dialogue datasets. This is because ECM directly incorporates the style information into its model architecture to force the generation of stylistic expressions. However, as shown in the quality scores, this behavior also undermines the quality of the generated responses. Therefore, the overall performance of ECM is not optimal as shown in the results of the ranking metric.



From the experiment results, we observe that removing retrieved information (PS w/o R) from the proposed model causes a drastic drop on the quality score. This demonstrates that the retrieved information is indispensable for the model to generate a stylistic response and maintain a high response quality. In addition, comparing with GPT2-FT baseline, the ablated model (PS w/o R) shows similar content quality and much stronger stylistic expression, which is gained from the model architectural design and the new training strategy. 


\begin{figure}[tb] 
	\centering    
	\setlength{\abovecaptionskip}{3pt}
	\includegraphics[width=0.35\textwidth]{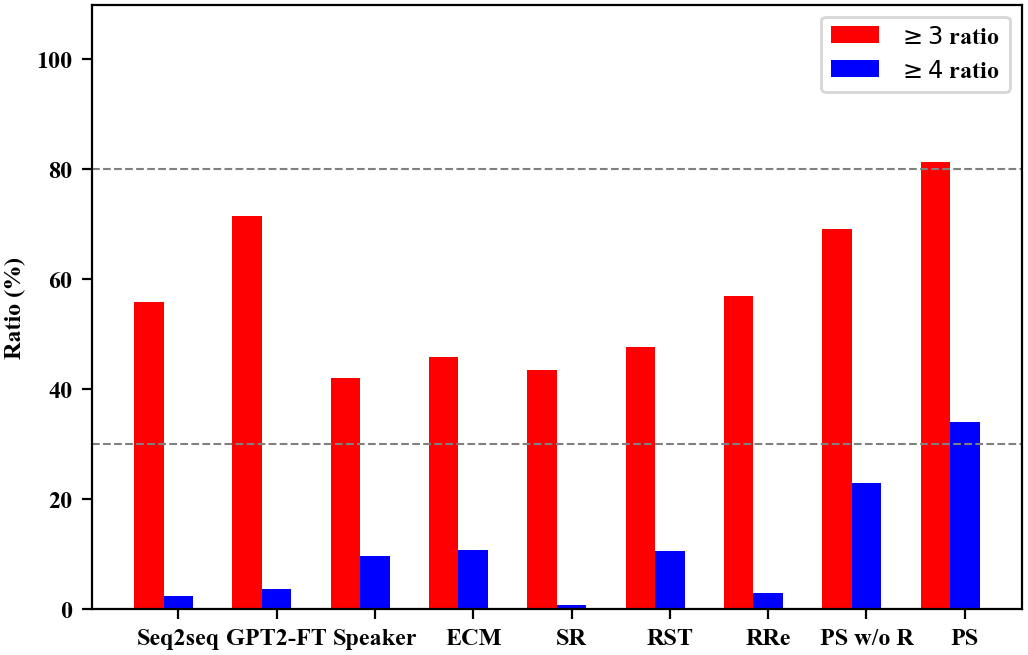}
	\caption{Balance between Quality and Style: The \textit{$\geq3$-ratio} means the ratio of responses whose both scores are greater or equal to $3$; \textit{$\geq4$-ratio} means the ratio of responses whose both scores are greater or equal to $4$.}
	\label{fig:quality_style_balance}
\end{figure}

\subsection{Further Analysis}
We present further discussions and empirical analysis of the proposed approach.
\subsubsection{Balance between Quality and Style}
In practice, a satisfactory stylistic dialogue system should express the desired style on the premise of the response quality. Based on the criterion of human evaluation metric, $3$ is the marginal score of acceptance. So we deem a response as marginally acceptable by actual users when both quality and style expression scores are greater or equal to $3$. On the other hand, $4$ is the score that well satisfies the users, so responses with both scores greater or equal to $4$ are deemed as satisfying to actual users. 

The ratios of both scores $\geq3$ and $\geq4$ are shown in Figure \ref{fig:quality_style_balance}, from which we can see that the proposed approach outperforms all other approaches on \textit{$\geq3$-ratio} and \textit{$\geq4$-ratio}. The proposed model best balances the trade-off between the response quality and style expression and therefore generating most acceptable and satisfying responses.

\begin{figure}[tb] 
 	\centering    
 	\setlength{\abovecaptionskip}{3pt}
 	\includegraphics[width=0.35\textwidth]{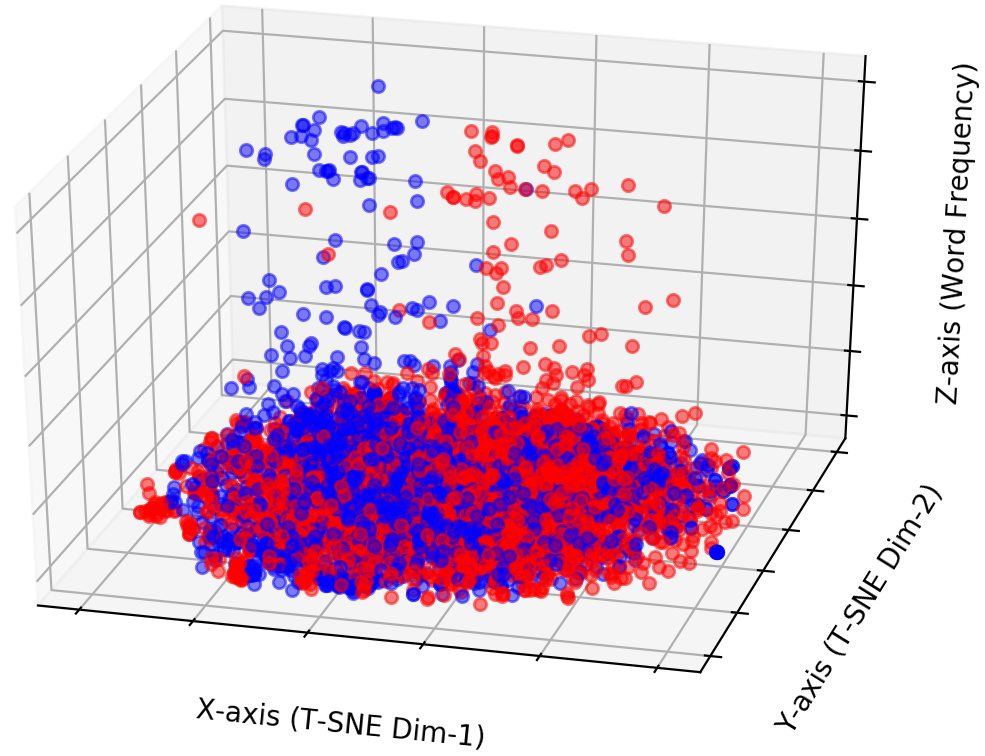}
 	\caption{Blue and red dots represent the words in gender-specific and emotion-specific dataset. Each word $w_d$ is denoted as $(x_d, y_d, z_d)$ where $(x_d, y_d)$ is T-SNE representation of its pretrained Glove embeddings \cite{DBLP:conf/emnlp/PenningtonSM14} and $z_d$ is the word frequency in the corresponding dataset. A notable distribution discrepancy between two domains can be observed.}
 	\label{fig:distribution}
 \end{figure}
 
\begin{table*}[tb] 
	\centering    
	\setlength{\abovecaptionskip}{3pt}
	\includegraphics[width=0.93\textwidth]{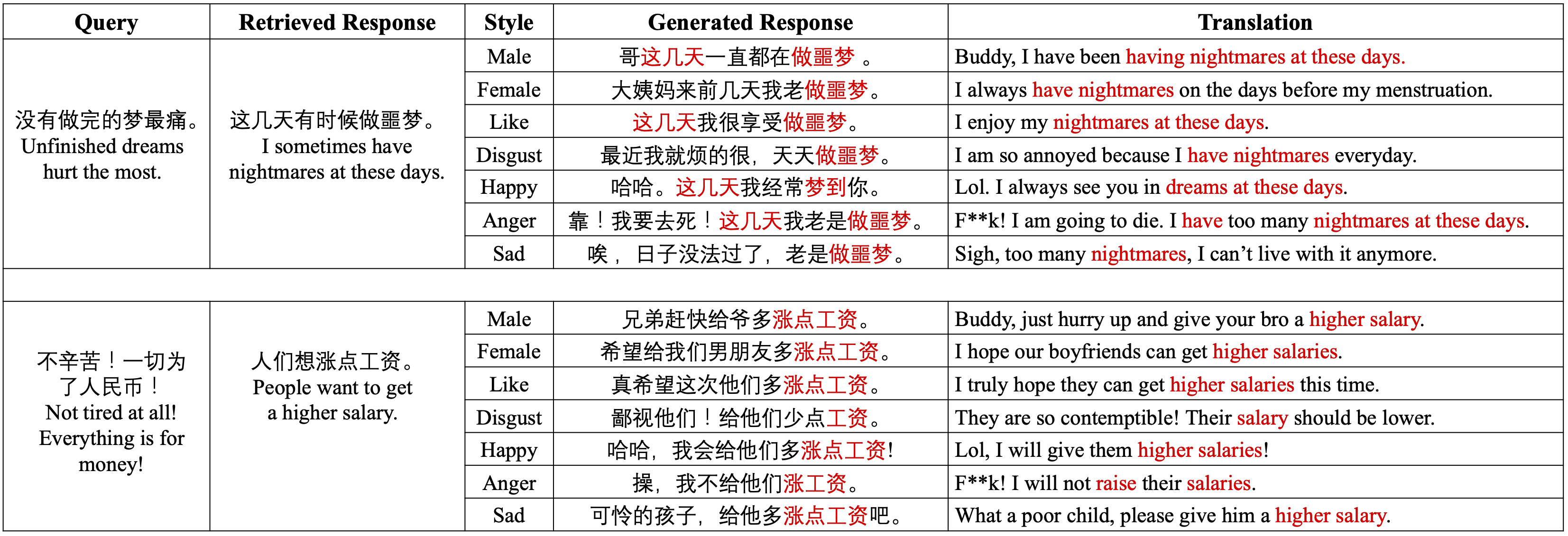}
	\caption{Examples of generated responses with different gender and emotion styles. The words in red color are the informative details that the model extracts from the retrieved response.}
	\label{tb:chinese_example}
\end{table*}

\begin{table*}[tb] 
	\centering    
	\setlength{\abovecaptionskip}{3pt}
	\includegraphics[width=0.93\textwidth]{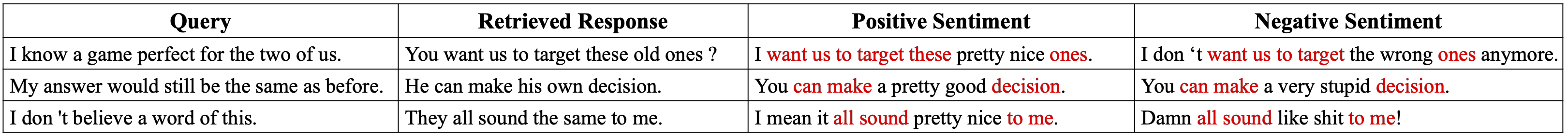}
	\caption{Examples of generated responses with different sentiments.}
	\label{tb:english_example}
\end{table*}

\subsubsection{Cross-Domain Evaluation}
To evaluate the robustness of different approaches, we further analyze their performances when there is a notable difference between the data distribution of the training and testing set. Specifically, we use the models trained on gender-specific dataset to conduct inference on the test set of emotion-specific dataset and vise versa, which is regarded as domain variation. In Figure \ref{fig:distribution}, we show the data distributions of these two datasets from which we can observe a notable distribution discrepancy. For evaluation, all results are evaluated with the same metrics as in the previous experiments. The averages response quality scores before and after domain variation are shown in Figure \ref{fig:cross_quality}\footnote{The concrete numerical results of cross-domain evaluation are shown in the supplementary material.}. For a direct comparison, the in-domain performance of each model can be found in Table \ref{tb:gender-result} and \ref{tb:emotion-result}.  


\begin{figure}[tb] 
	\centering    
	\setlength{\abovecaptionskip}{3pt}
	\includegraphics[width=0.35\textwidth]{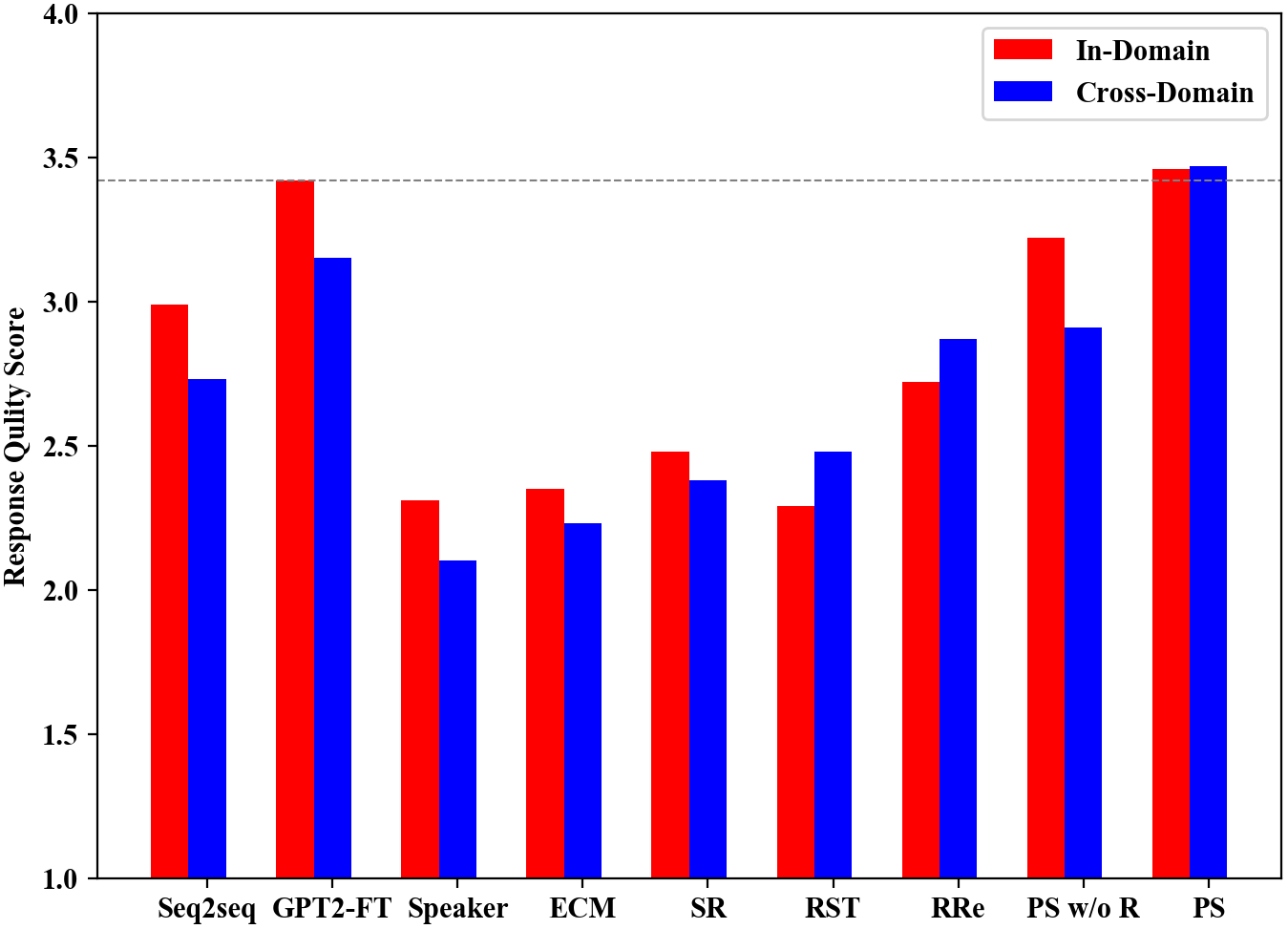}
	\caption{In-domain and cross-domain evaluations on the quality of generated responses. The red column represents the averaged quality score on in-domain test set, and the blue column denotes the averaged quality score after domain variation}
	\label{fig:cross_quality}
\end{figure}

As shown in Figure \ref{fig:cross_quality}, some of the strong baselines exhibit a drastic drop in response quality after domain variation such as GPT2-FT and PS w/o R. In contrast, the PS model successfully maintains high response quality in spite of domain variation. The model seems to benefit from leveraging retrieved results to bridge the gap between the two different domains. This can also be observed in the results of RST and RRe which also use the retrieved results and get a even higher performance when facing domain variation.

\subsection{Case Study}
We present several examples of generated responses by the proposed PS approach. Table \ref{tb:chinese_example} shows responses with different gender and emotion styles, and Table \ref{tb:english_example} shows responses with different sentiments.
Examples in Table \ref{tb:chinese_example} show that the proposed approach is able to extract informative details such as \textit{``have nightmares''} and \textit{``higher salary''} that are relevant to the queries from the retrieved responses. By taking the desired style as input, the proposed model generates adequate and stylistic responses while producing the informative details. 
Examples in Table \ref{tb:english_example} also demonstrate that the proposed model is able to generate responses with desired sentiments based on the informative details (e.g. \textit{``\_ want us to target \_ ones \_''}, \textit{``\_ can make \_ decision.''} and \textit{``\_ sound \_ to me \_''}) contained in the retrieved response.


\section{Conclusion}
In this work, we propose a novel PS framework to tackle the task of stylistic dialogue generation. Additionally, we propose a new stylistic response generator which works coherently with the proposed framework. We conduct extensive experiments on three benchmark datasets from two languages. Results of human and automatic evaluation show that the proposed approach outperforms many strong baselines by a substantial margin. 


\bibliography{acl2020}
\bibliographystyle{acl_natbib}
\end{document}


\clearpage
\onecolumn
\section*{Supplementary Material}
\section{Dataset}
\subsection{Gender-Specific Dataset}
Since there is no off-the-shelf dialogue dataset which specifically considers the gender style of the responses. To facilitate future research in this area, we manually collected over 4.5 million query-response pairs from popular Chinese social media platforms, including Tieba, Zhidao, Douban and Weibo. From the collected dataset,  we randomly select a subset of 200,000 query-response pairs and recruit five professional annotators from an annotation company to annotate the gender style contained in the responses. For each query-response pair, each annotator is asked to assign one of the three labels (\textit{Male}, \textit{Female} or \textit{Neutral}) to the response. Labels are selected if at least four out of five annotators agree.

The final annotated subset contains 5,184 male instances and 10,710 female instances. To proper balance the data distribution, we then randomly select 15,000 neutral instances. Therefore, the final gender style response classification dataset contains 5,184 male, 10,710 female and 15,000 neutral instances. Some examples of the annotated dataset are shown in Table \ref{tb:gender30k}.
\begin{table}[h] 
	\centering    
	\setlength{\abovecaptionskip}{3pt}
	\includegraphics[width=0.5\textwidth]{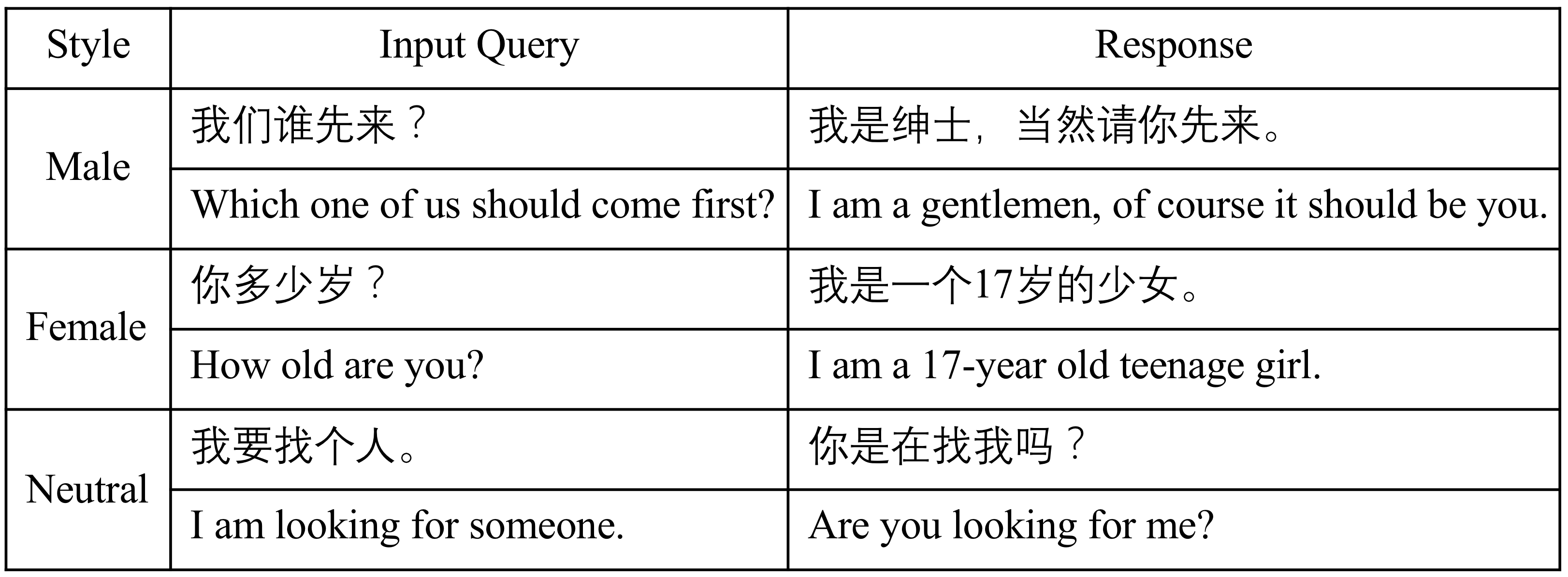}
	\caption{Examples of gender response classification dataset : Both Chinese and translated versions are provided.}
	\label{tb:gender30k}
\end{table}

After building the gender classification dataset, we train a gender classifier based on BERT \cite{DBLP:conf/naacl/DevlinCLT19} which achieves a 91.7\% accuracy. Then we use the trained classifier to automatically annotate the collected 4.5 million query-response pairs to get a large gender-specific dialogue dataset. The data statistic of the gender-style dialogue dataset is shown in Table \ref{tb:gen_stat}. To facilitate future research in this area, both the gender classification dataset and the large gender-specific dialogue dataset will be made publicly available.

\begin{table}[h]
    \footnotesize
	\setlength{\abovecaptionskip}{3pt}
	\renewcommand\arraystretch{1.1}
	\centering  
    \begin{center}
    \scalebox{1.0}{
	\begin{tabular}{|c|c|c|c|}
	    \hline
		Queries&\multicolumn{2}{|c|}{4,579,712}&Percentage(\%)\\
		\hline
        \multirow{3}{*}{Responses}&Male&68,350&1.49\%\\\cline{2-4}
        &Female&206,654&4.51\%\\\cline{2-4}
        &Neutral&4,304,708&94.00\%\\
        \hline
	\end{tabular}}
	\end{center}
	\caption{Data Statistic of Gender-Specific Dialogue Dataset}
	\label{tb:gen_stat}
\end{table}

\subsection{Emotion-Specific Dataset}
In addition to the gender-specific dataset, we use a publicly available emotion-specific dataset \cite{DBLP:conf/aaai/ZhouHZZL18} which is also written in Chinese. This dataset contains responses with six different emotions including \textit{Like}, \textit{Disgust}, \textit{Happy}, \textit{Anger}, \textit{Sad} and \textit{Other}. We refer the readers to the original paper for more details of this dataset.

\subsection{Sentiment-Sepcific Dataset}
To evaluate the proposed model's performance across different languages, we also conduct experiments on a English sentiment-specific dialogue dataset. Specifically, we build the sentiment-specific dataset on the basis of the publicly available OpenSubtitles\footnote{We use the data preprocessed by \citet{DBLP:conf/acl/LiGBSGD16}.} dataset \cite{DBLP:conf/lrec/LisonT16}. 

To train a sentiment classifier, we resort to the SemEval-2017 Subtask A dataset \cite{rosenthal-etal-2017-semeval}, which consists of twitter instances with different sentiments (\textit{Positive}, \textit{Negative} and \textit{Neutral}). The sentiment classifier is also constructed with BERT which achieves 71.4\% classification accuracy. The trained classifier is further used to annotate the OpenSubtitles dataset and the resulting data statistic is shown in Table \ref{tb:senti_stat}.




\section{Experiments}
\subsection{Chinese Large-Scale Language Model Pre-training}
Because there is no off-the-shelf pre-trained word-level language model in Chinese, we manually pre-trained one ourselves. Specifically, we first collected a large-scale Chinese corpus from popular Chinese News sites, including Sina, Baidu, Tencent, Toutiao, BBC China and New York Times China. We preprocess the acquired corpus with PKUSEG \cite{pkuseg} tokenizer to create a word-level corpus. After filtering out invalid contents (e.g. URLs), the resulting corpus contains over 7.6 million sentences and over 350 million words. 

Then we build the transformer-based language model following the same configuration as the one of \citet{OpenAI_GPT}. We refer the readers to the original paper for more details. We pre-train our language model for 10 epochs with 4 GeForce GTX 1080 Ti GPUs.

\begin{table}[tb]
    \footnotesize
	\setlength{\abovecaptionskip}{3pt}
	\renewcommand\arraystretch{1.1}
	\centering  
    \begin{center}
    \scalebox{1.0}{
	\begin{tabular}{|c|c|c|c|}
	    \hline
		Queries&\multicolumn{2}{|c|}{26,265,224}&Percentage(\%)\\
		\hline
        \multirow{3}{*}{Responses}&Positive&4,275,978&16.28\%\\\cline{2-4}
        &Negative&6,282,641&23.92\%\\\cline{2-4}
        &Neutral&15,706,605&59.80\%\\
        \hline
	\end{tabular}}
	\end{center}
	\caption{Data Statistic of Sentiment-Specific Dialogue Dataset}
	\label{tb:senti_stat}
\end{table}

\subsection{Implementation Details}
For experiments on different datasets, we limit the vocabulary size as 20,000. To optimize the proposed model, we use Adam \cite{DBLP:journals/corr/KingmaB14} optimizer with a batch size of 64 and learning rate of 2e-5. For each dataset, the model is trained for 3 epochs. 

At inference stage, for simplicity, we build the retrieval system based on the training corpus. Specifically, given a new input query, the Jaccard similarity \cite{lipkus1999proof} is measured between the new input query and queries contained in the training corpus. Then we select the response of the most similar query in the training corpus as the retrieved response.

\section{Cross-Domain Evaluation}
In this section, we present evaluation results of different models when facing the domain variation. To this end, we use the model trained on gender-specific dataset to conduct inference on the test set of the emotion-specific dataset and vice versa. The results are evaluated by the annotators following the same protocol as the one in previous experiments. The numerical results of cross-domain evaluation are shown in Table \ref{tb:gender-result} and \ref{tb:emotion-result}.

\begin{table*}[h]
    \footnotesize
	\setlength{\abovecaptionskip}{3pt}
	\renewcommand\arraystretch{1.1}
	\centering  
    \begin{center}
    \scalebox{0.8}{
	\begin{tabular}{|c|c|c|c|c|c|c|c|c|c|c|}
		\hline
		\multirow{ 2}{*}{Style}&\multirow{ 2}{*}{Metrics}&\multicolumn{4}{|c|}{Generative}&\multicolumn{3}{|c|}{Retrieval-Based}&\multicolumn{2}{|c|}{Ours}\\
		\cline{3-11}
		&&Seq2seq&GPT2-FT&Speaker&ECM&SR&RST&RRe&PS w/o R&PS\\\hline
		\hline
        \multirow{3}{*}{Male}& Quality$\uparrow$&2.72&3.13&2.17&2.43&2.26&2.22&2.49&3.02&\textbf{3.40}\\
        &Style Expression$\uparrow$&3.01&2.99&3.90&3.72&2.99&3.28&2.99&3.61&\textbf{3.79}\\
        &Ranking$\downarrow$&3.56&3.29&3.56&3.36&4.41&3.88&4.10&2.17&\textbf{1.64}\\
        \hline
        \multirow{3}{*}{Female}& Quality$\uparrow$&2.74&3.15&2.76&2.54&2.28&2.20&2.54&2.93&\textbf{3.23}\\
        &Style Expression$\uparrow$&2.99&3.03&2.97&3.22&3.01&3.20&3.04&3.56&\textbf{3.91}\\
        &Ranking$\downarrow$&3.26&2.90&3.27&3.33&3.98&3.65&3.63&2.09&\textbf{1.60}\\
        \hline
        \multirow{3}{*}{Overall}& Quality$\uparrow$&2.73&3.14&2.46&2.49&2.27&2.21&2.52&2.98&\textbf{3.31}\\
        &Style Expression$\uparrow$&3.00&3.01&3.44&3.48&3.00&3.24&2.99&3.59&\textbf{3.85}\\
        &Ranking$\downarrow$&3.41&3.09&3.42&3.34&4.20&3.76&3.87&2.13&\textbf{1.62}\\
        \hline
	\end{tabular}}
	\end{center}
	\caption{\underline{Cross-Domain Evaluation} Results on Gender-Specific Dialogue Generation: ($\uparrow$ means the higher the better and $\downarrow$ means the lower the better . The best results of each metric are presented in bond font.) Sign tests on human evaluation scores show that our full model significantly outperforms other models with p-value $<$ 0.05.}
	\label{tb:gender-result}
\end{table*}

\begin{table*}[h]
    \footnotesize
	\setlength{\abovecaptionskip}{3pt}
	\renewcommand\arraystretch{1.1}
	\centering  
    \begin{center}
    \scalebox{0.8}{
	\begin{tabular}{|c|c|c|c|c|c|c|c|c|c|c|}
		\hline
		\multirow{ 2}{*}{Style}&\multirow{ 2}{*}{Metrics}&\multicolumn{4}{|c|}{Generative}&\multicolumn{3}{|c|}{Retrieval-Based}&\multicolumn{2}{|c|}{Ours}\\
		\cline{3-11}
		&&Seq2seq&GPT2-FT&Speaker&ECM&SR&RST&RRe&PS w/o R&PS\\\hline
		\hline
        \multirow{3}{*}{Like}& Quality$\uparrow$&2.75&3.17&2.11&2.30&2.44&2.57&3.03&3.10&\textbf{3.61}\\
        &Style Expression$\uparrow$&2.74&2.92&3.89${\dagger}$&\textbf{4.32}${\dagger}$&3.03&3.19&2.88&3.76${\dagger}$&3.71\\
        &Ranking$\downarrow$&4.26&4.03&4.08&3.19&4.37&3.76&3.69&2.27&\textbf{1.81}\\
        \hline
        \multirow{3}{*}{Disgust}& Quality$\uparrow$&2.76&3.18&1.99&2.10&2.46&2.52&3.18&2.88&\textbf{3.61}\\
        &Style Expression$\uparrow$&2.63&2.67&\textbf{3.96}${\dagger}$&3.78${\dagger}$&2.90&3.10&2.93&3.50${\dagger}$&3.46\\
        &Ranking$\downarrow$&4.30&4.18&3.93&4.00&4.23&3.74&3.21&2.56&\textbf{1.79}\\
        \hline
        \multirow{3}{*}{Happy}& Quality$\uparrow$&2.70&3.15&1.78&2.23&2.41&2.63&2.93&3.04&\textbf{3.50}\\
        &Style Expression$\uparrow$&3.97&3.41&4.84${\dagger}$&\textbf{4.95}${\dagger}$&3.03&4.63&2.91&4.80${\dagger}$&4.70\\
        &Ranking$\downarrow$&3.88&4.48&4.39&3.42&5.41&2.95&4.76&1.98&\textbf{1.57}\\
        \hline
        \multirow{3}{*}{Anger}& Quality$\uparrow$&2.72&3.17&1.83&1.84&2.41&2.61&3.09&2.54&\textbf{3.51}\\
        &Style Expression$\uparrow$&1.80&2.35&3.94${\dagger}$&\textbf{4.03}${\dagger}$&2.94&3.46&2.91&3.92${\dagger}$&3.81\\
        &Ranking$\downarrow$&5.48&4.76&4.37&4.20&4.40&3.34&3.48&2.64&\textbf{1.71}\\
        \hline
        \multirow{3}{*}{Sad}& Quality$\uparrow$&2.69&3.15&2.04&2.19&2.39&2.60&2.84&2.85&\textbf{3.45}\\
        &Style Expression$\uparrow$&1.99&2.44&\textbf{3.62}${\dagger}$&3.32&2.86&3.14&2.96&3.56&\textbf{3.61}\\
        &Ranking$\downarrow$&4.96&4.24&4.15&4.06&4.10&3.34&3.45&2.31&\textbf{1.67}\\
        \hline
        \multirow{3}{*}{Overall}& Quality$\uparrow$&2.73&3.16&1.95&2.13&2.42&2.59&3.01&2.88&\textbf{3.54}\\
        &Style Expression$\uparrow$&2.63&2.76&4.05${\dagger}$&\textbf{4.08}${\dagger}$&2.95&3.50&2.92&3.91${\dagger}$&3.86\\
        &Ranking$\downarrow$&4.59&4.34&4.18&3.77&4.50&3.43&3.72&2.35&\textbf{1.71}\\
        \hline
	\end{tabular}}
	\end{center}
	\caption{\underline{Cross-Domain Evaluation} Results on Emotional-Specific Dialogue Generation: ($\uparrow$ means the higher the better and $\downarrow$ means the lower the better . The best results of each metric are presented in bond font.) Sign tests on human evaluation scores show that our full model significantly outperforms other models with p-value $<$ 0.05 with the only exception marked by ${\dagger}$.}
	\label{tb:emotion-result}
\end{table*}

\begin{figure}[tb] 
	\centering    
	\setlength{\abovecaptionskip}{3pt}
	\includegraphics[width=0.4\textwidth]{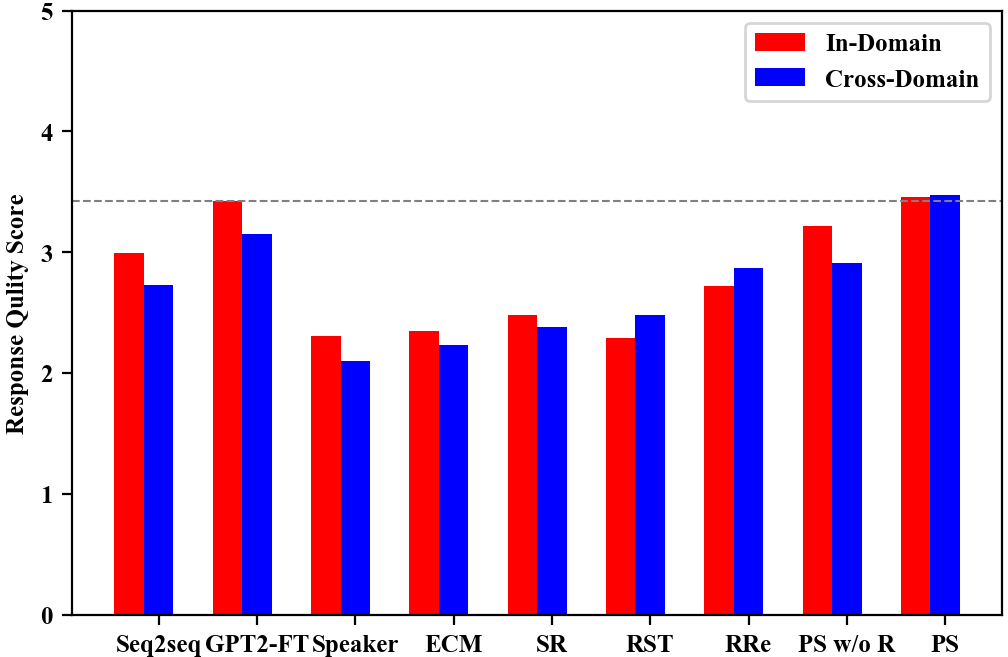}
	\caption{In-domain and cross-domain evaluations on the quality of the generated responses. The red column represents the averaged quality score on in-domain test set, and the blue column denotes the averaged quality score after domain variation.}
	\label{fig:quality}
\end{figure}

\begin{figure}[tb!] 
	\centering    
	\setlength{\abovecaptionskip}{3pt}
	\includegraphics[width=0.4\textwidth]{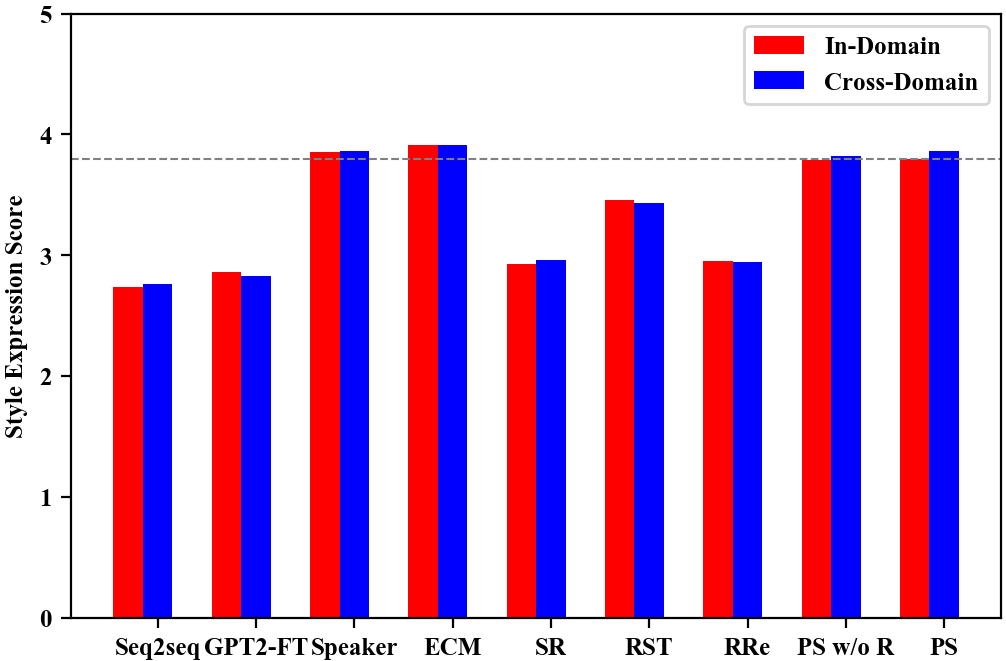}
	\caption{In-domain and cross-domain evaluations on the style expression of the generated responses. The red column represents the averaged ranking on in-domain test set, and the blue column denotes the averaged style expression score after domain variation.}
	\label{fig:style}
\end{figure}

\begin{figure}[h] 
	\centering    
	\setlength{\abovecaptionskip}{3pt}
	\includegraphics[width=0.4\textwidth]{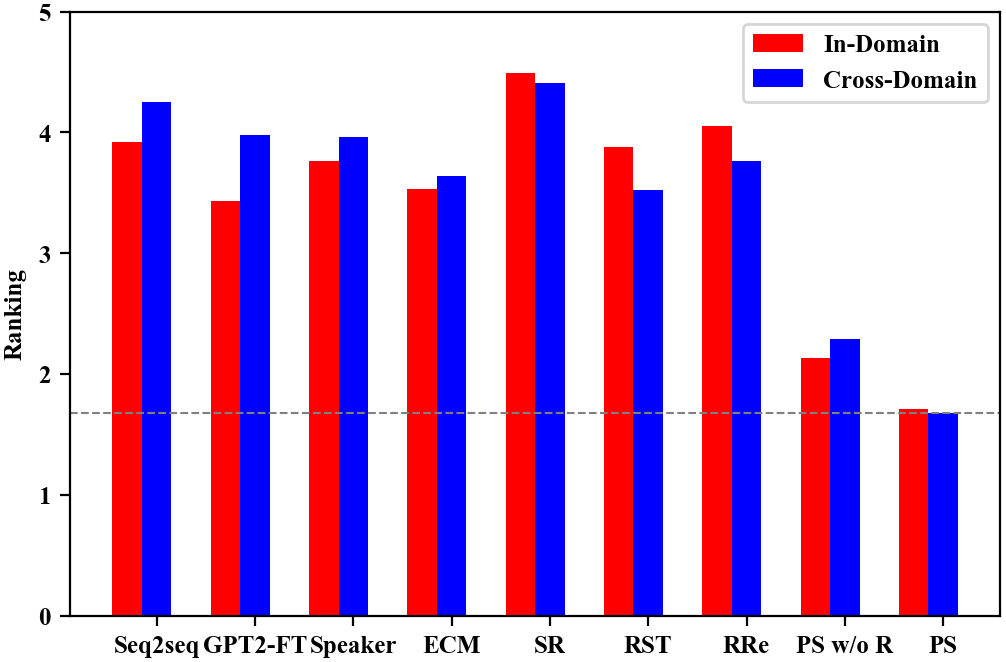}
	\caption{In-domain and cross-domain evaluations on the ranking of generated responses. The red column represents the averaged ranking on in-domain test set, and the blue column denotes the averaged ranking after domain variation.}
	\label{fig:ranking}
\end{figure}

We also compare the results of both in-domain and cross-domain evaluations.
The results for quality, style expression and ranking are shown in Figure \ref{fig:quality}, \ref{fig:style} and \ref{fig:ranking} respectively. 

Firstly, in Figure \ref{fig:quality}, drastic drop after domain-variation can be found on some strong baselines like GPT2-FT and PS w/o R. In contrast, the PS model successfully maintains high response quality after domain variation. It is benefit from the leveraging of retrieved results which helps to bridge the gap between the two different domains. The same effect can also be observed in the results of RST and RRe which also use the retrieved results and get a even higher performance when facing domain variation.

Secondly, looking at style expression performance in Figure \ref{fig:style}, we can see that there is not obvious difference between the results of in-domain and cross-domain evaluations. Our analysis is that, to generate responses with desired language style, the model could simply generate the characteristic expressions for that language style without considering the input query. Therefore, the domain variation actually poses little effect on the performance of style expression.

Finally, by jointly considering the quality and style expression, from Figure \ref{fig:ranking} we can see that the proposed model achieves best ranking for both in-domain and cross-domain evaluation. Therefore, it is safe to say that the proposed model is the best and the most robust one among all approaches.

\bibliography{acl2020}
\bibliographystyle{acl_natbib}